  \documentclass[letterpaper]{article} 
  \usepackage{aaai2027}  
  \usepackage[hyphens]{url}  
  \usepackage{graphicx} 
  \usepackage{natbib}  
  \usepackage{caption} 
  \usepackage{amsmath}
  \usepackage{amssymb}
  \usepackage{latexsym}
  \usepackage{algorithm}
  \usepackage{tabularx}
  \usepackage{booktabs}
  \usepackage{array}
\nocopyright
  \title{When Vision Becomes Text: Visual Token Pruning via Cross-Modal Residual Guidance in VLMs}

  \makeatletter
  \renewcommand\@fnsymbol[1]{\ensuremath{\ifcase#1\or \dagger\or \ddagger\or
    \mathsection\or \mathparagraph\or \|\or *\or {**}\or \dagger\dagger
    \or \ddagger\ddagger \else\@ctrerr\fi}}
  \makeatother

  \author{
  Congyang Ou\textsuperscript{\rm 1,2},
  Ruike Song\textsuperscript{\rm 1},
  Yang Zhou\textsuperscript{\rm 1},
  Libo Sun\textsuperscript{\rm 3},
  Haokui Zhang\textsuperscript{\rm 1,}\thanks{Corresponding author.},
  Zhenbo Luo\textsuperscript{\rm 2}
  }

  \affiliations{
  \textsuperscript{\rm 1}Northwestern Polytechnical University\\
  \textsuperscript{\rm 2}MiLM Plus, Xiaomi Inc.\\
  \textsuperscript{\rm 3}Chongqing University\\
  }

  \usepackage{colortbl}
  \usepackage{multirow}
  \definecolor{HeaderGray}{RGB}{245,245,245}
  \definecolor{BestBlue}{RGB}{0,114,178}
  \definecolor{OursBlue}{RGB}{227,242,253}

\begin{document}
  \maketitle

\begin{abstract}

Abundant visual information strengthens vision-language model (VLM) perception, yet massive visual tokens raise inference costs. Existing visual token pruning methods rely on similarity-based guidance, which exploits pairwise text-vision and vision-vision token correlations for compression. However, such methods only capture local layer-level signals and overlook the whole inference process in VLM. 
In this paper, we revisit VLM inference and present a new efficient guidance scheme that complements similarity-based guidance. 
In particular, we identify a key observation: as LLM layers deepen, text tokens continuously aggregate visual information via self-attention and progressively absorb partial visual content into textual representations.
To quantify this phenomenon, we propose Cross Modal Absorption (CMA) from a geometric representation perspective to measure how much visual information is absorbed by text, revealing that more visual tokens in deeper layers can be approximately explained by the text subspace. 
We accordingly propose Cross Modal Residual (CMR). It projects visual tokens onto the text subspace via Tikhonov regularized least squares and exploits reconstruction residuals to quantify visual information that cannot be explained by text.
Finally, based on CMR, we present SIEVE, a training-free visual token compression method that combines CMR, text-attention relevance, and residual-space diversity to retain task-relevant and complementary tokens. Experiments on diverse VLM architectures verify the effectiveness of SIEVE. For instance,  on LLaVA-NeXT-7B, SIEVE keeps only $11.1\%$ of visual tokens while preserving $97.5\%$ of the original average performance, achieving $3.62\times$ prefill speedup, $2.49\times$ end-to-end speedup, and a $6.02\times$ KV-cache reduction.






\end{abstract}

\section{Introduction}\label{Introduction}

Vision Language Models (VLMs) have achieved remarkable progress in multimodal tasks such as visual question answering, image captioning, and document understanding. Representative models, including LLaVA~\cite{liu2023llava}, InternVL~\cite{chen2024internvl}, and Qwen-VL~\cite{bai2023qwen}, typically encode images into numerous visual tokens and feed them into large language models together with text tokens for joint reasoning. However, the increasing number of visual tokens substantially raises self-attention computation, leading to higher inference latency and computational cost. This overhead is further amplified in high-resolution images, multi-image inputs, and video understanding. Therefore, compressing visual tokens while preserving model capability is crucial for efficient VLM inference.

Existing visual token compression methods mainly select tokens based on current layer attention~\cite{chen2024FastV,zhang2025sparsevlm} or feature similarity~\cite{bolya2023tome,wen2025dart}: the former characterizes the relevance between visual tokens and the current textual context, whereas the latter measures the redundancy among visual tokens. Although these signals provide useful criteria, they only characterize the local state at a single layer and lack a holistic view of the entire multimodal reasoning process. To this end, we revisit the inference process of VLMs, rather than staying at the local state of a single layer, and observe that as LLM layers deepen, text tokens continuously aggregate visual information through self-attention, progressively integrating part of the visual content into the textual representations. 

\begin{figure}[t]
\centering
\includegraphics[width=0.45\textwidth,height=0.25\textwidth,keepaspectratio]{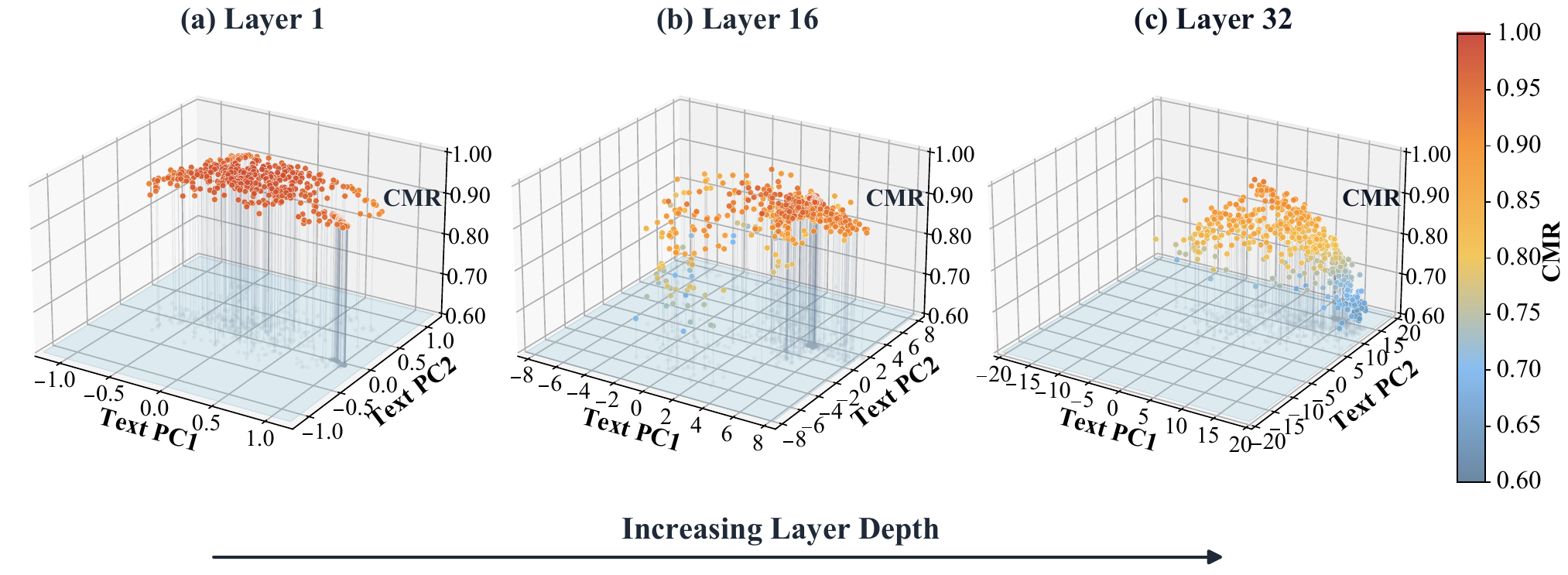}
\caption{3D visualization of visual tokens relative to the text subspace. Text tokens are centered and projected by PCA to obtain the first two principal directions, denoted as Text PC1 and Text PC2. Visual tokens are centered using the same text mean and projected onto these directions to obtain planar coordinates, while height and color indicate their CMR.
More results are provided in Appendix~A.}

\label{fig:cmr_3d_visualization}
\end{figure}

To quantify this phenomenon, we propose Cross Modal Absorption (CMA) from a geometric representation perspective, which measures at the layer level the overall extent to which visual token representations are absorbed by text, i.e., explainable by the text subspace. CMA characterizes the absorption trend at the layer level, whereas compression must operate on individual tokens. We therefore refine this perspective to the token level and propose Cross Modal Residual (CMR). Specifically, each visual token is projected onto the text subspace with Tikhonov regularization, and the reconstruction residual is used to quantify the visual information unexplained by text. A larger residual indicates that the token carries more unique visual information and thus has a higher retention necessity. This projection can be efficiently solved in closed form via regularized least squares, introducing only minor additional overhead. As shown in Figure~\ref{fig:cmr_3d_visualization}, visual token representations gradually move closer to the text subspace as layer depth increases, indicating that visual information does not always remain independent, but is progressively integrated into the text representation space.


Building on the above analysis, we propose \textbf{SIEVE}, a training-free visual token compression method that selects tokens from three complementary perspectives: CMR for visual distinctiveness and text-to-visual attention relevance for task relevance, so that retained tokens both carry unique visual information and align with the current textual context, together with diversity selection in the residual space to remove redundancy. Unlike de-duplication in the original feature space, the residual space has already removed the components explainable by text, so its directions more directly reflect the unique visual information of each token, enabling more precise identification of genuinely complementary tokens. SIEVE is training-free and plug-and-play, making it directly applicable to existing VLMs.

In summary, our contributions are threefold:
\begin{itemize}


\item We propose a novel pruning guidance criterion named CMR. Unlike mainstream token pruning approaches guided by token similarity, we revisit the model inference process and identify the phenomenon where visual information is gradually assimilated by textual information. We then quantify this process to develop CMR.


\item We propose SIEVE, a training-free visual token compression method that jointly selects tokens using CMR, attention relevance, and residual-space diversity selection, balancing visual distinctiveness, task relevance, and complementarity. 

\item Extensive experiments across three VLM architectures and eight benchmarks demonstrate that the proposed method significantly reduces computational overhead and memory, delivering practical inference acceleration.


\end{itemize}

\section{Related Work}\label{Related Work}
\subsection{Multimodal Large Language Models}
VLMs perform multimodal reasoning by feeding visual and text tokens into LLMs.
Early systems use cross-attention or query-based designs, e.g.,
Flamingo~\cite{alayrac2022flamingo}, BLIP-2~\cite{li2023blip},
Kosmos-1~\cite{huang2023kosmos}, and PaLM-E~\cite{driess2023palme}, followed by
instruction-tuned models such as MiniGPT-4~\cite{zhu2023minigpt4},
mPLUG-Owl~\cite{ye2023mplugowl}, LLaVA~\cite{liu2023llava}, and
InstructBLIP~\cite{dai2023instructblip}. Recent high-resolution, general-purpose
VLMs such as Qwen-VL~\cite{bai2023qwen}, InternVL~\cite{chen2024internvl},
Gemini~\cite{team2023gemini}, LLaVA-NeXT~\cite{liu2024llavanext}, and
Qwen2.5-VL~\cite{bai2025qwen2} further improve OCR, grounding, and visual
reasoning, and extend to video understanding~\cite{maaz2024video-chatgpt,lin2024video-llava,li2024llavaonevision,li2024videochat2}.
However, high-resolution images, multi-image inputs, and long videos rapidly
inflate visual-token counts, incurring heavy self-attention and KV-cache
overhead; low-loss visual token compression is thus crucial for their efficient
inference.
\subsection{Visual Token Compression for VLMs}
Existing methods mainly fall into two categories. \textit{Attention-score
pruning}~\cite{chen2024FastV,zhang2025sparsevlm,xing2024pyramiddrop,liu2025dynamicllava}
removes low-importance tokens without training, but attention scores are biased
and unreliable indicators of true importance. \textit{Similarity-based
reduction}~\cite{bolya2023tome,wen2025dart,yang2025visionzip,shang2024prumerge,li2024tokenpacker}
merges similar tokens to preserve visual coverage, yet similarity need not
correspond to text-query relevance. Other methods use diversity- and
coverage-based selection~\cite{alvar2025divprune,zhang2025cdpruner,deng2025scope,holov},
adaptive token dropping~\cite{chen2026v2drop}, cross-modal information
flow~\cite{lin2025vtw,huang2025topv} or objective-driven~\cite{zhang2026trio}. These criteria, however, capture only a
single-layer local state and lack a holistic view of the multimodal reasoning
process. We therefore revisit the VLM inference process and find that, as layers deepen, text tokens progressively absorb visual information, reducing the independent contribution of some visual tokens. This suggests that token retention should focus on the residual information unexplained by text.

\begin{figure}[!t]
    \centering
    \includegraphics[width=0.37\textwidth]{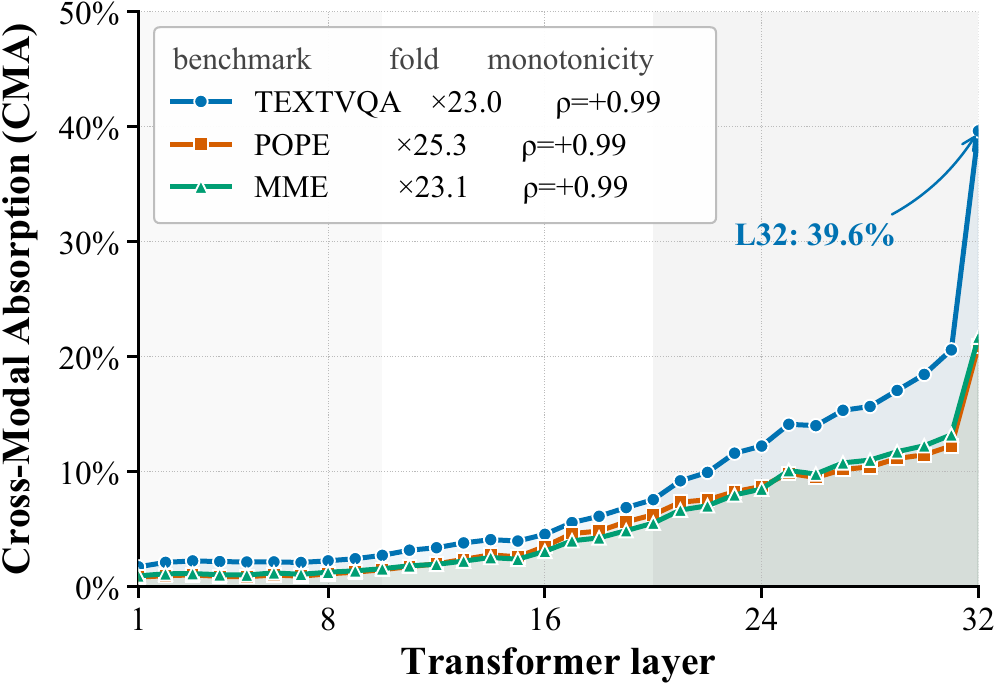}
    \caption{
Layer-wise evolution of cross-modal absorption in LLaVA-1.5-7B.
Cross-modal absorption (CMA) increases across Transformer layers.
    }
    \label{fig:cma_kurt}
\end{figure}

\section{Analysis}
\label{sec:analysis}


Differing from these works, we revisit the VLM inference process and investigate the evolution of visual tokens and text tokens throughout this procedure. Figure \ref{fig:cma_kurt} presents our observation.

\begin{figure*}
    \centering
    \includegraphics[width=0.75\textwidth,height=0.45\textwidth,keepaspectratio]{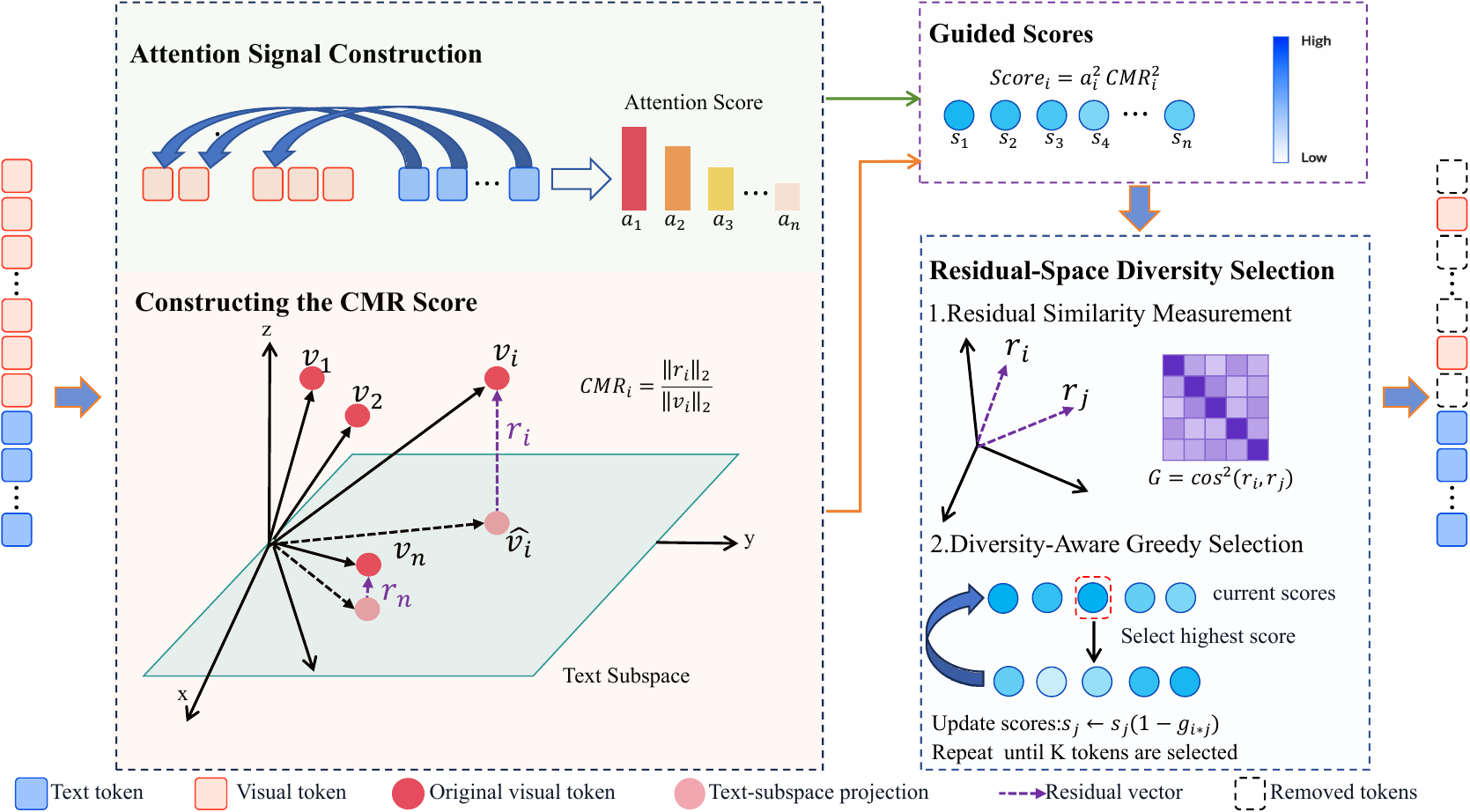}
    \caption{
    Overview of SIEVE. SIEVE measures each visual token’s residual information with respect to the text subspace via CMR, combines it with attention relevance, and performs residual-space diversity selection to retain informative and non-redundant visual tokens for the compressed input sequence.
    }
    \label{fig:overall}
\end{figure*}





We analyze all 32 layers of LLaVA-1.5-7B on three benchmarks, TextVQA~\cite{singh2019textvqa}, POPE~\cite{li2023pope}, and MME~\cite{fu2025mme}, covering 16,284 samples. The results show that an increasing portion of visual-token representations can be explained by the text subspace, indicating that visual information is progressively integrated into the text representation space during forward propagation.

\subsection{Measuring Cross Modal Absorption}
\label{sec:measuring_cma}
To quantify this phenomenon, we define \textbf{Cross Modal Absorption (CMA)} to measure how much of the visual token representation at a given layer can be explained by the text subspace. The CMA at layer $l$ is defined as
\begin{equation}
\mathrm{CMA}=
1 -
\frac{
\left\|
V_c - \hat{V}_c
\right\|_F^2
}{
\left\|
V_c
\right\|_F^2
}.
\end{equation}
Here, $V_c$ denotes the matrix of all centered visual tokens at layer $l$ (rather than an individual token); $\hat{V}_c$ its projection onto the text subspace; and $||\cdot||_F$ the Frobenius norm. CMA is a layer-level metric: a larger value means more visual information can be explained by the text subspace, i.e., stronger cross modal absorption.

\subsection{Visual Tokens Are Progressively Absorbed by the Text Subspace}
\label{sec:cma_results}

As shown in Figure~\ref{fig:cma_kurt}, CMA consistently increases from shallow to deep layers across all three benchmarks: from $0.008$ to $0.209$ on POPE, from $0.009$ to $0.217$ on MME, and from $0.017$ to $0.396$ on TextVQA. Its Spearman correlation with layer depth reaches $\rho=+0.99$ on all three, indicating a highly consistent monotonic trend.

Thus visual tokens do not remain fully independent throughout the network, but become increasingly explainable by the text subspace as depth grows. This reflects cross modal information flow in autoregressive VLMs: under causal attention, text tokens continuously aggregate visual information, so the text subspace comes to contain more visual-related directions. \textbf{It also reveals cross modal redundancy in deep visual tokens: a token already well explained by the text subspace may not need to be fully preserved}. We can therefore measure, at the token level, the relationship between each visual token and the text subspace, keeping information that is hard to explain by text while removing redundant tokens with little text-unexplained residual. Accordingly, the next section introduces the token-level Cross-Modal Residual (CMR) and the CMR-based visual token compression algorithm, SIEVE.

Note that the absolute value of CMA should not be read directly on the $[0,1]$ scale. Since the number of text tokens is usually around $N_t \approx 30$, far smaller than the hidden dimension $D=4096$, a random subspace can theoretically explain only about $N_t/D \approx 0.007$ of a high-dimensional representation. Thus the final-layer CMA of $0.209$ on POPE is already about $29\times$ the random baseline, indicating that the explanatory power of the text subspace over visual representations is not random but gradually emerges as a cross-modal representation pattern in deeper layers. Details are provided in Appendix B.

\section{Method}\label{Method}
Since CMA operates at the layer level while compression acts on individual tokens, we refine it to the token level and propose \textbf{SIEVE}, a training-free visual token compression method that selects tokens from three complementary perspectives: CMR, text-attention relevance, and residual-space diversity, as shown in Figure~\ref{fig:overall}

\subsection{Constructing the CMR Score}

SIEVE first constructs a Cross Modal Residual (CMR) score to measure the residual information in each visual token that cannot be explained by the text subspace: a lower score means higher redundancy, a higher one more distinctive visual information. At the $l$-th pruning layer, let $T \in \mathbb{R}^{N_t \times D}$ and $V \in \mathbb{R}^{N_v \times D}$ denote the text- and visual-token representation matrices, where $D$ is the hidden dimension. SIEVE first centers both modalities using the mean text representation:
\begin{equation}
\bar{t}
=
\frac{1}{N_t}
\sum_{j=1}^{N_t}
T_j,
\quad
T_c
=
T-\bar{t},
\quad
V_c
=
V-\bar{t}.
\end{equation}
Here $\bar{t}$ is the mean text representation, and $T_c$, $V_c$ are the centered text and visual matrices. Using the text center as reference reduces the background components shared between the two modalities and highlights each visual token's unique deviation from the text subspace.

For the $i$-th centered visual token $v_{c,i}$, SIEVE reconstructs it as a linear combination of text tokens, using a Tikhonov-regularized least-squares formulation for stability:
\begin{equation}
b_i
=
\arg\min_{b_i \in \mathbb{R}^{1 \times N_t}}
\left\|
v_{c,i}
-
b_iT_c
\right\|_2^2
+
\lambda
\left\|
b_i
\right\|_2^2 .
\end{equation}
Here $b_i$ are the combination coefficients of $v_{c,i}$ over text tokens and $\lambda$ is the regularization coefficient. The first term measures the reconstruction error from the text subspace, while the second constrains the coefficient magnitude to prevent unstable solutions caused by low-energy directions.

To adaptively determine $\lambda$, SIEVE constructs the text Gram matrix:
\begin{equation}
G
=
T_cT_c^\top .
\end{equation}
Here $G \in \mathbb{R}^{N_t \times N_t}$ characterizes the energy distribution of the text subspace, providing the basis for setting $\lambda$. Sorting its eigenvalues in descending order as $\sigma_1^2 \geq \cdots \geq \sigma_{N_t}^2$ and given an energy ratio threshold $\eta$, SIEVE selects the smallest $k$ satisfying
\begin{equation}
\sum_{j=1}^{k}
\sigma_j^2
\geq
\eta
\sum_{j=1}^{N_t}
\sigma_j^2 .
\end{equation}
Here $\eta$ is the energy proportion to be covered and $k$ is the number of dominant directions needed to reach it, delimiting the effective dimensionality of the text subspace. The regularization coefficient is then set as
\begin{equation}
\lambda
=
\sigma_k^2
+
\epsilon ,
\end{equation}
where $\epsilon$ is a small constant for numerical stability. Setting $\lambda$ to the $k$-th eigenvalue lets the dominant directions contribute fully to reconstruction while suppressing instability from low-energy directions. This yields the closed-form solution and the reconstruction:
\begin{equation}
b_i
=
v_{c,i}T_c^\top
\left(
G+\lambda I
\right)^{-1},
\qquad
\hat{v}_{c,i}
=
b_iT_c .
\end{equation}
Here $I$ is the identity matrix and $\hat{v}_{c,i}$ is the visual token reconstructed from the text subspace; the closed-form solution avoids iterative optimization and adds only negligible overhead. Finally, SIEVE defines the CMR score as the ratio between the reconstruction residual and the original centered representation:
\begin{equation}
\mathrm{CMR}_i
=
\frac{
\left\|
v_{c,i}
-
\hat{v}_{c,i}
\right\|_2
}{
\left\|
v_{c,i}
\right\|_2
}.
\end{equation}
Normalizing the residual by the original norm removes the effect of differing token magnitudes. A lower $\mathrm{CMR}_i$ means the token is well reconstructed by the text subspace and can be preferentially pruned, while a higher $\mathrm{CMR}_i$ means it preserves a larger text-unexplained residual and should be preferentially retained.
\begin{table*}[t]
\centering
\small
\setlength{\tabcolsep}{7pt}

\begin{tabular}{l|l|cccccccc|c}
\toprule
\textbf{Methods} & \textbf{Venue}
& \textbf{GQA} & \textbf{MMB} & \textbf{MMB-cn} & \textbf{MME}
& \textbf{POPE} & \textbf{SQA} & \textbf{VQAv2} & \textbf{TextVQA}
& \textbf{Average} \\
\midrule

\textbf{LLaVA-1.5-7B}
&
& \multicolumn{8}{c|}{\textbf{Upper Bound, 576 Tokens}}
& \\
Vanilla & --
& 61.9 & 64.7 & 58.1 & 1862 & 85.9 & 69.5 & 78.5 & 58.2
& \textbf{100\%} \\

\midrule

\textbf{LLaVA-1.5-7B}
&
& \multicolumn{8}{c|}{\textbf{Retain 192 Tokens (33.3\%)}}
& \\
FastV & ECCV'24
& 52.7 & 61.2 & 57.0 & 1612 & 64.8 & 67.3 & 67.1 & 52.5
& 89.0\% \\
SparseVLM & ICML'25
& 59.5 & 64.1 & 53.7 & 1787 & 85.3 & 68.7 & 75.6 & 57.8
& 97.2\% \\
DART & EMNLP'25
& 60.0 & 63.6 & 57.0 & 1856 & 82.8 & 69.8 & 76.7 & 57.4
& 98.3\% \\
VisionZip & CVPR'25
& 59.3 & 63.0 & 57.3 & 1782 & 85.3 & 68.9 & 76.8 & 57.3
& 97.8\% \\
HoloV & NeurIPS'25
& 59.0 & 65.4 & 58.0 & 1820 & 85.6 & 69.8 & 76.7 & 57.4
& 98.7\% \\
SCOPE & NeurIPS'25
& 60.1 & 63.6 & 56.8 & 1804 & 86.4 & 68.8 & 77.2 & 57.7
& 98.3\% \\
SpecFlow & ICML'26
& 58.3 & 65.8 & 56.1 & 1827 & 85.8 & 69.7 & 76.4 & 57.9
& 98.4\% \\

\textbf{SIEVE}  & \textbf{Ours}
& 60.6 & 64.6 & 58.4 & 1820 & 85.5 & 68.9 & 77.8 & 57.8
& \textbf{99.1\%} \\

\midrule

\textbf{LLaVA-1.5-7B}
&
& \multicolumn{8}{c|}{\textbf{Retain 128 Tokens (22.2\%)}}
& \\
FastV & ECCV'24
& 49.6 & 56.1 & 56.4 & 1490 & 59.6 & 60.2 & 61.8 & 50.6
& 83.2\% \\
SparseVLM & ICML'25
& 58.4 & 64.5 & 51.1 & 1746 & 85.0 & 68.6 & 73.8 & 56.7
& 95.6\% \\
DART & EMNLP'25
& 58.7 & 63.2 & 57.5 & 1840 & 80.1 & 69.1 & 75.9 & 56.4
& 97.0\% \\
VisionZip & CVPR'25
& 57.6 & 62.0 & 56.7 & 1761.7 & 83.2 & 68.9 & 75.6 & 56.8
& 96.4\% \\
HoloV & NeurIPS'25
& 57.7 & 63.9 & 56.5 & 1802 & 84.0 & 69.8 & 75.5 & 56.8
& 97.2\% \\
SCOPE & NeurIPS'25
& 59.7 & 62.5 & 56.9 & 1776 & 86.1 & 68.4 & 76.5 & 57.2
& 97.5\% \\
SpecFlow & ICML'26
& 57.6 & 64.3 & 55.9 & 1794 & 84.9 & 69.9 & 75.3 & 56.8
& 97.2\% \\

\textbf{SIEVE}  & \textbf{Ours}
& 59.8 & 64.0 & 57.0 & 1815 & 85.8 & 68.8 & 77.1 & 57.6
& \textbf{98.7\%} \\

\midrule

\textbf{LLaVA-1.5-7B}
&
& \multicolumn{8}{c|}{\textbf{Retain 64 Tokens (11.1\%)}}
& \\
FastV & ECCV'24
& 46.1 & 48.0 & 52.7 & 1256 & 48.0 & 51.1 & 55.0 & 47.8
& 74.0\% \\
SparseVLM & ICML'25
& 53.8 & 60.1 & 52.7 & 1589 & 77.5 & 69.8 & 68.2 & 53.4
& 90.6\% \\
DART & EMNLP'25
& 55.9 & 60.6 & 53.2 & 1765 & 73.9 & 69.8 & 72.4 & 54.4
& 92.8\% \\
VisionZip & CVPR'25
& 55.1 & 60.1 & 55.4 & 1690 & 77.0 & 69.0 & 72.4 & 55.5
& 93.0\% \\
HoloV & NeurIPS'25
& 55.3 & 63.3 & 55.1 & 1715 & 80.3 & 69.5 & 72.8 & 55.4
& 94.4\% \\
SCOPE & NeurIPS'25
& 58.3 & 61.7 & 54.4 & 1698 & 83.9 & 68.6 & 75.3 & 56.6
& 95.4\% \\
SpecFlow & ICML'26
& 55.3 & 63.7 & 54.3 & 1713 & 80.5 & 69.7 & 73.7 & 54.9
& 94.4\% \\

\textbf{SIEVE}  & \textbf{Ours}
& 58.0 & 62.2 & 55.2 & 1740 & 83.5 & 69.0 & 75.2 & 56.5
& \textbf{96.0\%} \\
\bottomrule
\end{tabular}
\caption{Performance comparison of different token reduction methods on LLaVA-1.5-7B.}
\label{tab:llava15_main}
\end{table*}
\subsection{Attention Signal Construction}

CMR measures the unique information in a visual token that cannot be explained by the text subspace, but geometric uniqueness does not necessarily indicate relevance to the current text context. Therefore, SIEVE further introduces an attention signal to determine whether each visual token is actually attended to by the text tokens. For the $i$-th visual token, its text relevance score is defined as
\begin{equation}
a_i
=
\sum_{h \in \mathcal{H}_{\mathrm{top}}}
\sum_{t \in \mathcal{T}}
\alpha_{t,i}^{h}.
\end{equation}
Here, $\mathcal{T}$ denotes the set of text tokens, and $\alpha_{t,i}^{h}$ denotes the attention weight from text token $t$ to visual token $i$ in the $h$-th attention head. To reduce the influence of noisy heads, SIEVE keeps only the top $\rho_h$ proportion of heads with the largest attention mass on the visual region, denoted as $\mathcal{H}_{\mathrm{top}}$. A larger $a_i$ indicates stronger relevance to the current textual context. 

SIEVE does not need to explicitly store the full attention matrix. Following the dual-flash attention strategy in SparseVLM~\cite{zhang2025sparsevlm}, SIEVE can compute the required attention signal while remaining compatible with FlashAttention, with details provided in Appendix C.

\begin{table*}[t]
\centering
\small
\setlength{\tabcolsep}{5.2pt}

\begin{tabular}{l|l|cccccccc|c}
\toprule
\textbf{Method} & \textbf{Venue}
& \textbf{GQA} & \textbf{MMBench} & \textbf{MMBench-cn} & \textbf{MME}
& \textbf{POPE} & \textbf{SQA} & \textbf{VQAv2} & \textbf{TextVQA}
& \textbf{Avg} \\
\midrule

\textbf{LLaVA-NeXT-7B}
&
& \multicolumn{8}{c|}{\textbf{Upper Bound, 2880 Tokens}}
& \\
Vanilla & --
& 64.2 & 67.4 & 60.6 & 1851 & 86.5 & 70.1 & 81.8 & 61.4
& 100\% \\
\midrule

\textbf{LLaVA-NeXT-7B}
&
& \multicolumn{8}{c|}{\textbf{Retain 320 Tokens}}
& \\
FastV & ECCV'24
& 55.9 & 61.6 & 51.9 & 1661 & 71.7 & 62.8 & 71.9 & 55.7
& 88.1\% \\
SparseVLM & ICML'25
& 56.1 & 60.6 & 54.5 & 1533 & 82.4 & 66.1 & 71.5 & 58.4
& 90.3\% \\
VisionZip & CVPR'25
& 59.3 & 63.1 & 55.6 & 1702 & 82.1 & 67.3 & 76.2 & 58.9
& 93.7\% \\
DART & EMNLP'25
& 61.7 & 65.3 & 58.2 & 1710 & 84.1 & 68.4 & 79.1 & 58.7
& 96.1\% \\
HoloV & NeurIPS'25
& 61.7 & 65.3 & 57.5 & 1738 & 83.9 & 68.9 & 79.5 & 58.7
& 96.2\% \\
SpecFlow & ICML'26
& 62.5 & 66.7 & 56.8 & 1707 & 85.0 & 68.6 & 80.1 & 59.5
& 96.6\% \\

\textbf{SIEVE} & \textbf{Ours}
& 60.9 & 66.5 & 59.6 & 1819 & 85.7 & 68.3 & 78.7 & 59.6
& \textbf{97.5\%} \\
\bottomrule
\end{tabular}%
\caption{Performance comparison of different token reduction methods on LLaVA-NeXT-7B.}
\label{tab:llava16_main}
\end{table*}

\begin{table*}[t]
\centering
\small
\setlength{\tabcolsep}{7.1pt}

\begin{tabular}{l|l|ccccccc|c}
\toprule
\textbf{Method} & \textbf{Venue}
& \textbf{MMBench} & \textbf{MME} & \textbf{POPE} & \textbf{SQA}
& \textbf{MMBench-cn} & \textbf{GQA} & \textbf{TextVQA} & \textbf{Avg} \\
\midrule

\textbf{Qwen2.5-VL-7B} &
& \multicolumn{7}{c|}{\textbf{Upper Bound, 1296 Tokens}} & \\
Vanilla & --
& 83.4 & 2295 & 87.1 & 88.7 & 81.4 & 61.0 & 77.2 & 100.0\% \\
\midrule

\textbf{Qwen2.5-VL-7B} &
& \multicolumn{7}{c|}{\textbf{Retain 33.3\% Tokens}} & \\
FastV & ECCV'24
& 79.5 & 2306 & 86.9 & 85.6 & 77.9 & 59.5 & 75.0 & 97.5\% \\
HoloV & NeurIPS'25
& 81.2 & 2288.9 & 86.1 & 88.7 & 80.0 & 60.2 & 74.4 & 98.5\% \\

\textbf{SIEVE}  & \textbf{Ours}
& 82.5 & 2319 & 88.0 & 88.2 & 81.3 & 60.8 & 75.8 & \textbf{99.7\%} \\
\midrule

\textbf{Qwen2.5-VL-7B} &
& \multicolumn{7}{c|}{\textbf{Retain 22.2\% Tokens}} & \\
FastV & ECCV'24
& 76.5 & 2243 & 85.1 & 84.8 & 75.2 & 57.8 & 73.5 & 95.0\% \\
HoloV & NeurIPS'25
& 80.5 & 2268.8 & 85.2 & 88.2 & 79.6 & 59.5 & 71.7 & 97.2\% \\

\textbf{SIEVE}  & \textbf{Ours}
& 81.6 & 2313 & 87.3 & 87.5 & 80.2 & 60.3 & 75.1 & \textbf{98.9\%} \\
\midrule

\textbf{Qwen2.5-VL-7B} &
& \multicolumn{7}{c|}{\textbf{Retain 11.1\% Tokens}} & \\
FastV & ECCV'24
& 71.2 & 2061 & 78.6 & 80.6 & 68.2 & 52.4 & 67.0 & 87.5\% \\
HoloV & NeurIPS'25
& 78.2 & 2246.5 & 83.7 & 87.6 & 77.1 & 57.5 & 65.7 & 94.4\% \\

\textbf{SIEVE}  & \textbf{Ours}
& 80.0 & 2273 & 86.0 & 86.6 & 77.6 & 58.9 & 72.5 & \textbf{96.7\%} \\
\bottomrule
\end{tabular}%
\caption{Performance comparison of different token reduction methods on Qwen2.5-VL-7B.}
\label{tab:qwen2.5}
\end{table*}

\subsection{Guided Scores}

SIEVE initializes each token's integrated score as
\begin{equation}
\mathrm{Score}_i = a_i^2 \, \mathrm{CMR}_i^2,
\end{equation}
where $a_i$ is the text-relevance (attention) score and $\mathrm{CMR}_i$ is the residual ratio unexplained by the text subspace; the product keeps attention from overlooking residual information and CMR from ignoring textual relevance.

\subsection{Residual-Space Diversity Selection}

Given the guided scores defined above, SIEVE aims to retain visual tokens that are both individually informative and mutually complementary. To this end, it performs residual-space diversity selection in two stages: it first measures the redundancy between visual tokens in the residual space, and then greedily selects tokens while progressively suppressing redundant candidates.

\begin{enumerate}
    \item \textbf{Residual Similarity Measurement.}
    Selecting the top-$K$ tokens by integrated score alone may retain tokens with similar residual directions, causing visual redundancy. To avoid this, SIEVE measures token redundancy in the \emph{residual space} rather than the original space: the residual removes the large shared component that otherwise dominates the cosine similarity, so comparisons reflect each token's genuinely complementary information. 
    For the $i$-th centered visual token $v_{c,i}$ with text-subspace reconstruction $\hat{v}_{c,i}$, the residual is
    \begin{equation}
    r_i = v_{c,i} - \hat{v}_{c,i},
    \end{equation}
    i.e., the part the text subspace cannot reconstruct. Redundancy between two tokens is measured by the squared cosine similarity of their residuals,
    \begin{equation}
    g_{ij}
    =
    \left(
    \frac{r_i^\top r_j}
    {\left\|r_i\right\|_2\left\|r_j\right\|_2}
    \right)^2
    \in [0,1],
    \end{equation}
    where a larger $g_{ij}$ means more similar residual directions and thus higher redundancy.

    \item \textbf{Diversity-Aware Greedy Selection.}
    Based on the guided scores and residual similarities, SIEVE iteratively selects informative and non-redundant visual tokens. At each step, it selects
    \begin{equation}
    i^\ast
    =
    \arg\max_{i \notin \mathcal{S}}
    \mathrm{Score}_i,
    \end{equation}
    with $\mathcal{S}$ the set of selected tokens, and then discounts the remaining candidates by their residual similarity to $i^\ast$:
    \begin{equation}
    \mathrm{Score}_j
    \leftarrow
    \mathrm{Score}_j(1-g_{i^\ast j}),
    \quad j \notin \mathcal{S}.
    \end{equation}
    Since $g_{i^\ast j}\in[0,1]$ is a dimensionless quantity normalized by the residual magnitudes, the factor $1-g_{i^\ast j}$ is a scaling coefficient in $[0,1]$ that attenuates $\mathrm{Score}_j$ proportionally without altering its dimension. Candidates whose residual directions align with the selected token are thereby progressively suppressed, and the process continues until the budget $K$ is reached.
\end{enumerate}

\section{Experiments}\label{Experiments}
\subsection{Experimental Setup}
\paragraph{Models and Baselines.}
We evaluate SIEVE on three VLMs of diverse architectures, including LLaVA-1.5~\cite{liu2023llava}, LLaVA-NeXT~\cite{liu2024llavanext}, and Qwen2.5-VL~\cite{bai2025qwen2}.

\paragraph{Datasets. }
We report image-based results on eight standard benchmarks: GQA~\cite{hudson2019gqa}, MMBench and MMBench-CN~\cite{liu2024mmbench}, MME~\cite{fu2025mme}, POPE~\cite{li2023pope}, SQA~\cite{lu2022sqa}, VQAv2~\cite{goyal2017vqav2}, and TextVQA~\cite{singh2019textvqa}.
\subsection{Comparison Experimental Results}

\noindent\textbf{Results on LLaVA.}
Table~\ref{tab:llava15_main} reports the results on LLaVA-1.5-7B.
Here, Vanilla denotes the original model without visual-token reduction.
On LLaVA-1.5-7B, SIEVE achieves the best average performance among methods with complete results under all token budgets.
With 192, 128, and 64 retained visual tokens, SIEVE preserves 99.1\%, 98.7\%, and 96.0\% of the Vanilla baseline, respectively.
Notably, even when retaining only 64 visual tokens, SIEVE still surpasses SCOPE, the strongest baseline with complete results, by 0.6 percentage points on average, demonstrating its ability to preserve critical visual information under aggressive compression.

\noindent\textbf{Results on LLaVA-NeXT-7B}. Table \ref{tab:llava16_main} presents the corresponding comparison results. SIEVE likewise achieves the best reported average performance with complete results under the 320-token setting, preserving 97.5\% of the Vanilla baseline.
Compared with DART and HoloV, SIEVE improves the average performance by 1.4 and 1.3 percentage points, respectively.
These results demonstrate that SIEVE generalizes effectively from the standard LLaVA-1.5 architecture to the higher-resolution LLaVA-NeXT architecture, which contains substantially more visual tokens.

\noindent\textbf{Results on Qwen2.5-VL.}
Table~\ref{tab:qwen2.5} reports the results on Qwen2.5-VL-7B.
We fix the input resolution to $1008 \times 1008$, corresponding to 1,296 visual tokens per image.
Under token retention ratios of 33.3\%, 22.2\%, and 11.1\%, SIEVE achieves 99.7\%, 98.9\%, and 96.7\% average performance, respectively, consistently outperforming both FastV and HoloV.
Notably, under the most aggressive 11.1\% token budget, SIEVE surpasses FastV and HoloV by 9.2 and 2.3 percentage points, respectively, demonstrating its stronger ability to preserve critical visual information under severe compression.
These results show that SIEVE generalizes effectively beyond the LLaVA family and maintains consistent performance advantages on the Qwen2.5-VL architecture.

\begin{table}[t]
    \centering
    \small
    \setlength{\tabcolsep}{8pt}
    \renewcommand{\arraystretch}{0.95}
    \begin{tabular}{lccccc}
        \toprule
        $\eta \backslash H_{\mathrm{top}}$
        & 8 & 12 & 16 & 24 & 32 \\
        \midrule
        0.65 & 96.15 & 96.35 & 96.29 & 96.42 & 96.26 \\ 
        0.75 & 96.17 & \textbf{96.59} & 96.07 & 96.05 & 96.20 \\
        0.85 & 96.25 & 96.48 & 96.15 & 96.16 & 96.29 \\
        0.95 & 96.37 & 96.40 & 96.27 & 96.47 & 96.35 \\
        \bottomrule
    \end{tabular}
    \caption{
        Hyperparameter sensitivity of SIEVE on LLaVA-1.5-7B at
        11.1\% visual-token retention. 
    }
    \label{tab:hyper_sensitivity_main}
\end{table}

\subsection{Ablation Study and Analysis}
\paragraph{Hyperparameter Sensitivity.}

We further evaluate the sensitivity of SIEVE to the energy ratio $\eta$ and the number of selected image-attending heads $H_{\mathrm{top}}$, with detailed results provided in Appendix D.

Table~\ref{tab:hyper_sensitivity_main} shows part results. SIEVE achieves the best average retention of 96.59\% with $\eta=0.75$ and $H_{\mathrm{top}}=12$.
Other configurations obtain highly similar results, ranging from 96.05\% to 96.52\%, indicating that SIEVE is robust to hyperparameter choices and does not rely on delicate tuning.


\begin{table}
\centering
\small
\setlength{\tabcolsep}{5pt}
\begin{tabular}{llrr}
\toprule
Group & Variant & Avg. \%Full & $\Delta$Avg. \\
\midrule
Full & \textbf{SIEVE full} & \textbf{96.59} & 0.00 \\
\midrule
Score & Attention score only & 96.07 & -0.52 \\
Score & CMR score only & 96.00 & -0.59 \\
Score & (Attention + CMR) score & 96.20 & 0.39 \\
\midrule
Diversity & Top-$K$ selection & 90.13 & -6.46 \\
Diversity & Raw-space diversity & 95.70 & -0.89 \\
\bottomrule
\end{tabular}
\caption{
Component ablation of SIEVE on LLaVA-1.5-7B at 11.1\% visual token retention. Avg. \%Full reports average performance on SQA, TextVQA, POPE, and MME relative to the full-token baseline, while $\Delta$Avg. is measured against full SIEVE. Ablations cover scoring and diversity selection.
}
\label{tab:component_ablation_main}
\end{table}

\paragraph{Component Ablation.}
To assess the contribution of each component, we ablate SIEVE on LLaVA-1.5-7B under 11.1\% visual token retention from two aspects: scoring signals and diversity selection, as shown in Table~\ref{tab:component_ablation_main}.

For scoring, using only the attention score and only the CMR score achieves 96.07\% and 96.00\%, respectively, showing that the geometric residual captured by CMR is itself an effective selection signal rather than an auxiliary term of attention. Combining the two signals multiplicatively gives the full SIEVE result of 96.59\%, outperforming either signal alone. In contrast, additive fusion reaches only 96.20\%, lagging behind multiplicative fusion.

For diversity selection, direct top-$k$ selection drops sharply to 90.13\% ($-6.46$), indicating that selecting tokens solely by the initial scores tends to retain redundant tokens. Raw-space diversity recovers most of the performance to 95.70\% ($-0.89$), but still underperforms full SIEVE, suggesting that enforcing diversity in the residual space is more effective for suppressing redundant visual information.

\begin{figure}[!t]
    \centering
    \includegraphics[width=0.4\textwidth]{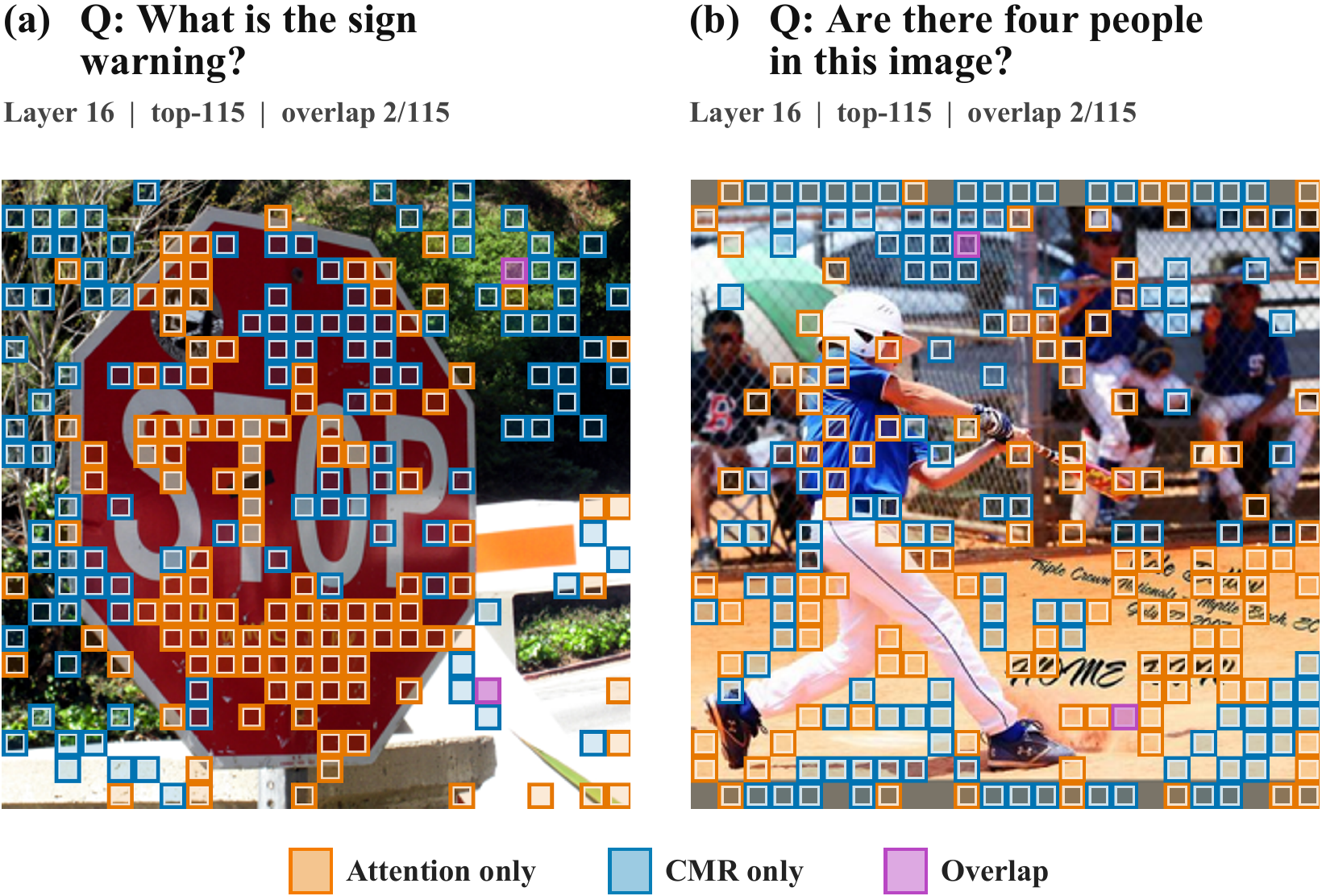}
    \caption{
Different selection between attention and CMR.
    }
    \label{fig:fusion}
\end{figure}

\paragraph{Effectiveness Analysis of the Fusion Strategy.}
Attention and CMR capture complementary aspects of token importance, namely semantic relevance and geometric residual, respectively. The former focuses on regions directly related to the query, whereas the latter highlights visual regions that cannot be well explained by the text subspace. As shown in Figure~\ref{fig:fusion}, although both respond to informative regions, the overlap between their selected tokens remains consistently low, with a low average overlap ratio across the four benchmarks (see Appendix~E for details). This indicates that the two signals are complementary rather than redundant: relying on either one alone inevitably misses tokens preferred by the other. Consequently, combining the two scores preserves a more complete set of informative visual tokens, leading to better performance than using either attention or CMR alone.

Furthermore, multiplicative fusion is better aligned with this objective than additive fusion. Additive fusion is sensitive to differences in score magnitude, allowing one signal to dominate the final ranking. In contrast, multiplicative fusion favors tokens that simultaneously exhibit semantic relevance and substantial residual information, while naturally suppressing background or noisy tokens that receive an abnormally high score from only one signal. As a result, additive fusion achieves only 96.20\%, lower than the 96.59\% obtained by multiplicative fusion.

\begin{table}
\centering
\small
\setlength{\tabcolsep}{3.2pt}

\begin{tabular}{ll|c|c|c|c}
\toprule
\textbf{Model} &
\textbf{Method} &
\begin{tabular}[c]{@{}c@{}}\textbf{Total}\\\textbf{Time}\end{tabular} &
\begin{tabular}[c]{@{}c@{}}\textbf{Prefill}\\\textbf{Time}\end{tabular} &
\begin{tabular}[c]{@{}c@{}}\textbf{KV Cache}\\\textbf{(MB)}\end{tabular} &
\begin{tabular}[c]{@{}c@{}}\textbf{POPE}\\\textbf{F1}\end{tabular} \\
\midrule

\multirow{4}{*}{\begin{tabular}[c]{@{}l@{}}LLaVA-\\NeXT-7B\end{tabular}}
& Full       & 46:11 & 37:21 & 1156 & 86.5 \\
& Random     & 15:33 & 7:20  & 192  & 79.1 \\
& SparseVLM  & 35:55 & 26:57 & 200  & 82.4 \\

& SIEVE       & 18:32 & 10:20 & 192  & 85.7 \\
\midrule

\multirow{4}{*}{\begin{tabular}[c]{@{}l@{}}LLaVA-\\1.5-7B\end{tabular}}
& Full       & 16:39 & 9:55 & 319 & 85.9 \\
& Random     & 10:49 & 4:10 & 63  & 75.8 \\
& SparseVLM  & 12:58 & 6:18 & 66  & 77.5 \\

& SIEVE       & 11:36 & 4:56 & 63  & 83.4 \\

\bottomrule
\end{tabular}%
\caption{
Efficiency comparison on POPE for LLaVA-NeXT-7B and LLaVA-1.5-7B.
All token reduction methods retain 11.1\% visual tokens.
}
\label{tab:efficiency_comparison}
\end{table}

\paragraph{Efficiency Analysis.}
SIEVE achieves a strong trade-off between efficiency and performance. As shown in Table~\ref{tab:efficiency_comparison}, compared with the full model, SIEVE achieves 2.49$\times$/1.44$\times$ total runtime speedups and 3.62$\times$/2.01$\times$ prefill speedups on LLaVA-NeXT/LLaVA-1.5, while reducing the KV cache from 1156MB/319MB to 192MB/63MB.
Even compared with Random pruning, which approximates an ideal fast baseline, SIEVE remains close in runtime, with only 19\%/7\% additional total time, while improving POPE F1 by 6.6/7.6 points.
Compared with SparseVLM, SIEVE is faster on both models in total runtime and prefill time while maintaining higher accuracy.

\section{Conclusion}\label{Conclusion}


In this work, we reveal cross-modal information absorption in VLMs and propose CMA for quantification. We design CMR via Tikhonov regularized subspace projection and develop the training-free SIEVE to compress visual tokens. Extensive evaluations over LLaVA, Qwen and LLaVA-NeXT verify that SIEVE substantially accelerates inference without obvious performance loss.

\bibliography{custom}

\newpage
\clearpage
\appendix
\section*{Appendix}
\setcounter{secnumdepth}{2}
\section{Additional Visualization of Cross-Modal Absorption}
\label{app:full_cmr_visualization}

\begin{figure*}[t]
    \centering
    \includegraphics[width=0.92\textwidth]{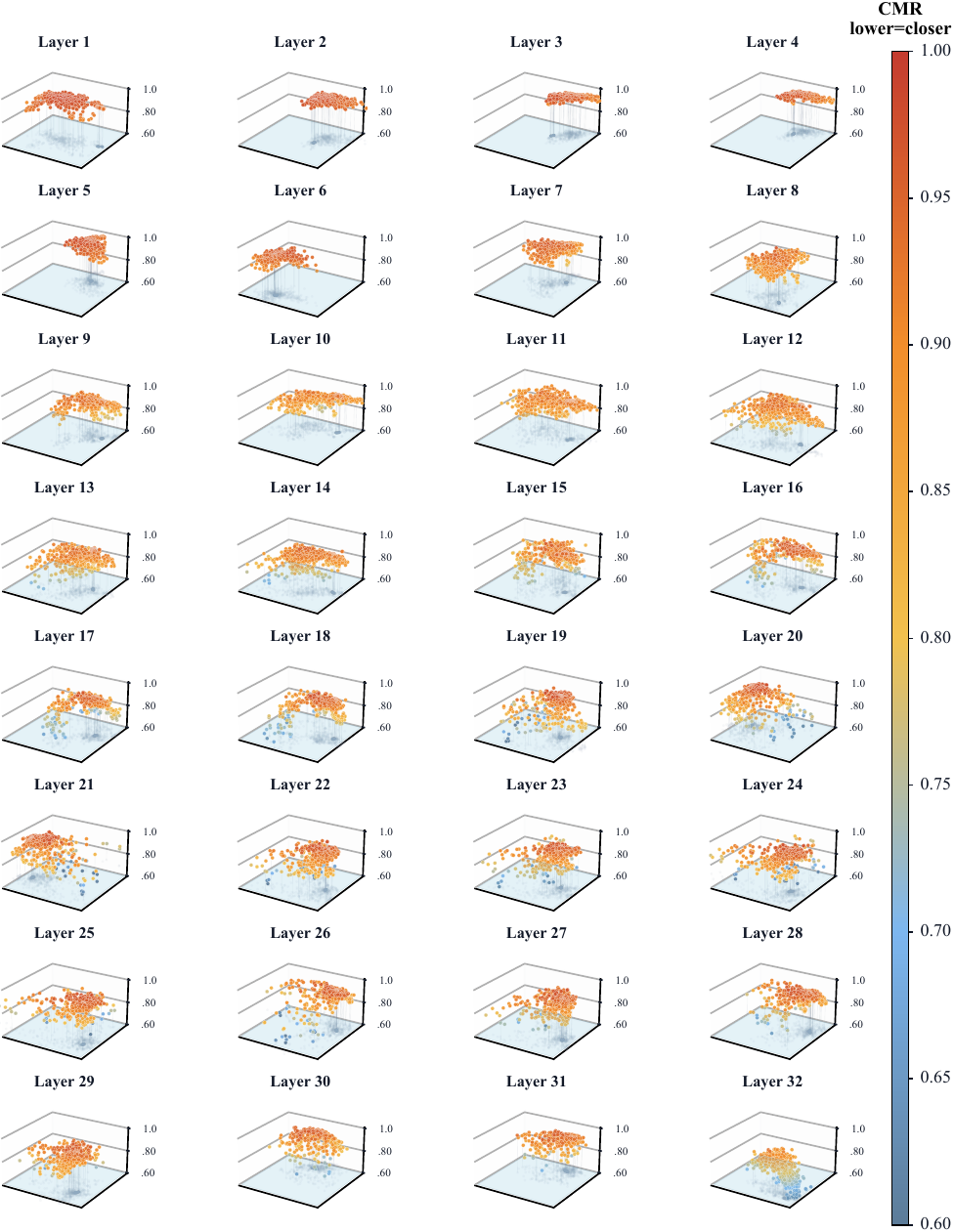}
    \caption{
    Full visualization of visual tokens progressively moving closer to the text subspace across all 32 layers.
    Each subplot corresponds to one Transformer layer.
    The point positions indicate the projections of visual tokens onto the text PCA plane, while the height and color represent CMR.
    A lower CMR indicates that the visual token is closer to the text subspace.
    }
    \label{fig:full_cmr_32_layers}
\end{figure*}

Figure~\ref{fig:full_cmr_32_layers} provides the complete visualization over all 32 layers, offering a more detailed view of how visual tokens evolve during layer-wise forward propagation.
Overall, visual tokens are progressively absorbed by the text subspace during layer-wise forward propagation.
As the layer depth increases, a large number of visual tokens gradually exhibit lower CMR and form a more concentrated distribution near the text PCA plane.
This suggests that the model continuously transforms visual representations into a subspace that becomes increasingly aligned with text representations in deeper layers.

Meanwhile, this absorption process does not occur uniformly across all visual tokens, but instead exhibits a highly non-uniform and sparse structure.
Most visual tokens gradually become redundant with the text subspace, meaning that their information can be well explained by text-related representations.
However, a small subset of tokens still maintains relatively high CMR even in deeper layers, indicating that they are not fully absorbed by the text subspace and may preserve unique visual information.
This phenomenon supports our key observation: visual information in multimodal models does not disappear synchronously as a whole, but is selectively absorbed by text representations layer by layer, where most tokens gradually become redundant while a few tokens remain complementary.

\section{Mathematical Derivation: The Scale of CMA and the Random Baseline}
\label{app:CMA_scale}

Section Analysis shows that the final-layer CMA value is usually around $0.2$--$0.4$. Although this value may appear small under the naive scale of $[0,1]$, it should be interpreted relative to the dimensionality of the text subspace. Since the number of text tokens is much smaller than the hidden dimension, the text subspace is intrinsically low-dimensional and cannot explain all visual information, even in the most favorable case.

This appendix provides a simple derivation of the effective scale of CMA. We first discuss the maximum CMA that a fixed low-dimensional subspace can achieve, then derive the random baseline when visual and text representations are independent, and finally compare these quantities with empirical observations.

\subsection{Upper Bound of CMA Under a Fixed Subspace Dimension}
\label{app:upper_bound}

Let $V_{\mathrm{vis}} \in \mathbb{R}^{N_v \times D}$ denote the visual-token matrix at a given layer, where $N_v$ is the number of visual tokens and $D$ is the hidden dimension. Let $U \in \mathbb{R}^{D \times k}$ be an orthonormal basis of a $k$-dimensional subspace, satisfying $U^\top U = I_k$. The projection of visual tokens onto this subspace is

\begin{equation}
\mathrm{proj}_U(V_{\mathrm{vis}})=V_{\mathrm{vis}}UU^\top .
\end{equation}

The corresponding CMA is defined as the fraction of visual information that can be explained by this subspace:

\begin{equation}
\begin{aligned}
\mathrm{CMA}(U)
&=
1-
\frac{
\|V_{\mathrm{vis}}-V_{\mathrm{vis}}UU^\top\|_F^2
}{
\|V_{\mathrm{vis}}\|_F^2
} \\
&=
\frac{
\|V_{\mathrm{vis}}U\|_F^2
}{
\|V_{\mathrm{vis}}\|_F^2
}.
\end{aligned}
\end{equation}

Here, the denominator denotes the total amount of visual information, while the numerator measures the part that lies in the subspace spanned by $U$.

To understand the maximum possible value of CMA, we decompose the visual representation into its principal directions. Let the singular values of $V_{\mathrm{vis}}$ be

\begin{equation}
\sigma_1 \geq \sigma_2 \geq \cdots \geq \sigma_R ,
\end{equation}
where $R=\min(N_v,D)$. The total visual information can then be written as

\begin{equation}
\|V_{\mathrm{vis}}\|_F^2
=
\sum_{r=1}^{R}\sigma_r^2 .
\end{equation}

Intuitively, $\sigma_r^2$ measures how much information is contained in the $r$-th principal direction. Therefore, if we can only use a $k$-dimensional subspace to explain the visual representation, the best possible choice is to align this subspace with the top-$k$ principal directions of $V_{\mathrm{vis}}$. These directions contain the largest amount of visual information.

Thus, for any $k$-dimensional subspace $U$, CMA is upper-bounded by

\begin{equation}
\mathrm{CMA}(U)
\leq
\frac{
\sigma_1^2+\sigma_2^2+\cdots+\sigma_k^2
}{
\sigma_1^2+\sigma_2^2+\cdots+\sigma_R^2
}.
\end{equation}

The equality holds only when the subspace $U$ is exactly aligned with the top-$k$ principal directions of $V_{\mathrm{vis}}$.

This result indicates that the upper bound of CMA is not $1.0$. Instead, it is determined by how much visual information is contained in the top-$k$ principal directions. In other words, a low-dimensional subspace is not able to explain all visual information.

In LLaVA-1.5-7B, the number of post-image text tokens is typically about $N_t \approx 30$, while the hidden dimension is $D=4096$. Therefore, the text subspace is very low-dimensional. Empirically, the top $30$ principal directions of visual representations usually account for about $40\%$--$70\%$ of the total visual information. As a result, even if the text subspace were perfectly aligned with the dominant visual directions, the theoretical upper bound of CMA would be around $0.4$--$0.7$, rather than $1.0$.

This explains why a final-layer CMA value of $0.2$--$0.4$ should not be considered small. It already occupies a substantial portion of the effective range that a low-dimensional text subspace can reach.

\subsection{Random Baseline Under Independent Visual and Text Representations}
\label{app:random_baseline}

We next consider the opposite case, where visual and text representations have no systematic alignment. In this case, the text subspace can be regarded as a random $k$-dimensional subspace in a $D$-dimensional space.

For a random $k$-dimensional subspace, the expected fraction of any vector that falls into this subspace is approximately

\begin{equation}
\frac{k}{D}.
\end{equation}

Therefore, when visual and text representations are independent, the expected CMA is

\begin{equation}
\mathbb{E}[\mathrm{CMA}]
=
\frac{k}{D}
\approx
\frac{N_t}{D},
\end{equation}

where $k=N_t$ corresponds to the effective dimension of the text subspace.

For LLaVA-1.5-7B, we have $N_t \approx 30$ and $D=4096$, which gives

\begin{equation}
\frac{N_t}{D}
\approx
\frac{30}{4096}
\approx
0.0073.
\end{equation}

Empirically, the layer-0 CMA values are $0.008$ on POPE, $0.009$ on MME, and $0.017$ on TextVQA. The values on POPE and MME are very close to the random baseline, while the value on TextVQA is slightly higher, possibly because this task already induces mild visual-text alignment in shallow layers.

Overall, the shallow-layer CMA values are of the same order as the random baseline. This suggests that visual and text representations are approximately independent in shallow layers, and that meaningful cross-modal absorption has not yet occurred.

\subsection{The Effective Scale for Interpreting CMA}
\label{app:scale}

Combining the upper-bound analysis and the random-baseline analysis, CMA should not be interpreted directly on the naive scale of $[0,1]$. Instead, its effective scale is approximately

\begin{equation}
\left[
\frac{N_t}{D},
\frac{
\sigma_1^2+\cdots+\sigma_{N_t}^2
}{
\sigma_1^2+\cdots+\sigma_R^2
}
\right]
\approx
[0.007,\;0.4\text{--}0.7].
\end{equation}

The left endpoint corresponds to the random baseline, where visual and text representations are nearly independent. The right endpoint corresponds to the ideal case, where the text subspace is perfectly aligned with the dominant visual directions.

Table~\ref{tab:CMA_scale} shows representative CMA values on POPE for LLaVA-1.5-7B.

\begin{table}[h]
\centering
\small
\setlength{\tabcolsep}{4pt}
\begin{tabular}{cccc}
\toprule
Layer & CMA & vs.\ Baseline & vs.\ Bound ($0.5$) \\
\midrule
0  & 0.008 & $\approx 1\times$ & $1.6\%$ \\
15 & 0.035 & $4.8\times$       & $7\%$   \\
25 & 0.095 & $13\times$        & $19\%$  \\
31 & 0.209 & $29\times$        & $42\%$  \\
\bottomrule
\end{tabular}
\caption{CMA at representative layers of LLaVA-1.5-7B on POPE, together with their relative positions on the effective scale.}
\label{tab:CMA_scale}
\end{table}

As shown in the table, the CMA value at layer $0$ is almost the same as the random baseline. In contrast, the final-layer CMA is about $29\times$ larger than the random baseline and reaches around $42\%$ of the theoretical upper bound.

Therefore, although the absolute CMA value at the final layer is only $0.209$, it is already far above random alignment and occupies a considerable portion of the attainable range of a low-dimensional text subspace. This supports our observation that visual tokens are progressively absorbed by the text subspace during the forward propagation of VLMs, revealing a systematic cross-modal phenomenon inside the model.

\section{FlashAttention-Compatible Attention Scoring}
\label{app:flash_compatibility}

SIEVE uses an attention signal to estimate how strongly post-image text tokens attend to visual tokens.
For the $h$-th attention head, let its causal attention matrix be
\begin{equation}
P^h =
\mathrm{Softmax}\left(
\frac{Q^h (K^h)^\top}{\sqrt{d_h}} + M
\right),
\end{equation}
where $M$ denotes the causal attention mask and $d_h$ is the head dimension.
Let $\mathcal{V}$ denote the set of visual-token positions and $T_{\mathrm{post}}$ denote the set of post-image text-token positions.
The attention score required by SIEVE is defined as
\begin{equation}
a_i =
\sum_{h\in H_{\mathrm{top}}}
\sum_{t\in T_{\mathrm{post}}}
P^h_{t,i},
\quad i\in\mathcal{V}.
\end{equation}
This score measures how much visual token $i$ is attended to by post-image text tokens at the current layer.
A direct implementation would require explicitly accessing the submatrix
$P^h_{T_{\mathrm{post}},\mathcal{V}}$.
However, FlashAttention does not materialize the full $N\times N$ attention matrix, but directly computes the attention output in a block-wise streaming manner:
\begin{equation}
O^h = P^h V^h .
\end{equation}
Therefore, explicitly reading $P^h_{t,i}$ is incompatible with the memory-efficient design of FlashAttention.

To avoid materializing $P^h$, SIEVE adopts a forward-only dual-flash aggregation strategy to compute the required text-to-visual attention statistics.
The idea is related to the dual-flash operation used in SparseVLM~\cite{zhang2025sparsevlm}, but the auxiliary value matrix is designed differently.
SparseVLM typically marks selected text raters in the auxiliary value matrix to obtain token-to-rater attention statistics.
In contrast, SIEVE requires the attention from post-image text queries to visual keys; therefore, the auxiliary value matrix should mark visual-token positions.

Let the visual-token positions be
\begin{equation}
\mathcal{V}=\{v_1,v_2,\ldots,v_{N_v}\}.
\end{equation}
We construct an auxiliary value matrix
$\widetilde V\in\mathbb{R}^{N\times N_v}$ whose $j$-th row is defined as
\begin{equation}
\widetilde V[j,:] =
\begin{cases}
e_k, & j=v_k,\ v_k\in\mathcal{V},\\
\mathbf{0}, & \text{otherwise},
\end{cases}
\end{equation}
where $e_k$ is the $k$-th standard basis vector.
Thus, each visual token is assigned an independent marker channel, while all non-visual tokens are assigned zero auxiliary values.
Using the same $Q^h$ and $K^h$, we compute the auxiliary attention output as
\begin{equation}
\widetilde O^h = P^h \widetilde V .
\end{equation}
Since $\widetilde V$ is nonzero only at visual-token positions, for any post-image text token $t\in T_{\mathrm{post}}$, we have
\begin{equation}
\widetilde O^h_t
=
\sum_{j=1}^{N} P^h_{t,j}\widetilde V[j,:]
=
\sum_{k=1}^{N_v} P^h_{t,v_k}e_k .
\end{equation}
Therefore, the $k$-th dimension of $\widetilde O^h_t$ is exactly the attention value $P^h_{t,v_k}$.
Aggregating over all post-image text tokens gives the text-to-visual attention received by each visual token under head $h$:
\begin{equation}
g^h_{v_k}
=
\sum_{t\in T_{\mathrm{post}}}
\widetilde O^h_{t,k}
=
\sum_{t\in T_{\mathrm{post}}}
P^h_{t,v_k}.
\end{equation}
This is equivalent to explicitly extracting $P^h_{T_{\mathrm{post}},\mathcal{V}}$ and summing over the text dimension, but it only requires computing $P^h\widetilde V$ and never materializes the full attention matrix.

After obtaining the per-head visual attention scores, SIEVE performs head selection.
For each head, we compute its total visual attention mass as
\begin{equation}
m_h =
\sum_{i\in\mathcal{V}} g_i^h .
\end{equation}
A larger $m_h$ indicates that the post-image text tokens in this head attend more strongly to the visual region, making the head more suitable for estimating visual-token relevance.
SIEVE selects the top image-attending heads:
\begin{equation}
H_{\mathrm{top}}
=
\mathrm{TopK}_h(m_h),
\end{equation}
and aggregates the final visual-token attention score over the selected heads:
\begin{equation}
a_i =
\sum_{h\in H_{\mathrm{top}}} g_i^h .
\end{equation}
The resulting attention score $a_i$ is then combined with the CMR score for residual-space diversity selection.

Importantly, this auxiliary computation does not modify the normal Transformer forward pass.
The main forward pass still computes the hidden states using the original value matrix:
\begin{equation}
O^h=P^hV^h.
\end{equation}
The auxiliary forward pass is only used to obtain the attention statistics required for pruning:
\begin{equation}
\widetilde O^h=P^h\widetilde V.
\end{equation}
Thus, SIEVE does not explicitly materialize the full attention matrix or alter the SIEVE FlashAttention computation.
By rewriting text-to-visual attention aggregation as $P^h\widetilde V$, SIEVE preserves post-image-text-guided attention scoring while remaining compatible with the streaming computation of FlashAttention.

\section{Ablation Study}

\subsection{Detailed Hyperparameter Sensitivity}
\label{app:hyper_sensitivity}

We provide a more detailed hyperparameter sensitivity analysis of SIEVE under the 11.1\% visual-token retention setting on LLaVA-1.5-7B.
The sweep covers two key hyperparameters: the energy ratio $\eta$ for Tikhonov-regularized text-subspace projection, and the number of selected image-attending heads $H_{\mathrm{top}}$ for attention-score aggregation.
For each configuration, we compute the average percentage over the full-token baseline across SQA, TextVQA, POPE, and MME.

Table~\ref{tab:head_topk_sensitivity} summarizes the effect of $H_{\mathrm{top}}$.
For each value of $H_{\mathrm{top}}$, we report the mean, best, and worst average performance over all tested energy ratios.
The results show that selecting too few heads leads to slightly lower performance, suggesting that overly aggressive head filtering may miss useful visual relevance signals.
The best mean performance is achieved when $H_{\mathrm{top}}=12$, indicating that a moderate number of image-attending heads provides a good trade-off between capturing visual relevance and filtering noisy heads.

\begin{table}[ht]
\centering
\small
\setlength{\tabcolsep}{6pt}
\begin{tabular}{cccc}
\toprule
$H_{\mathrm{top}}$ & Mean over $\eta$ & Best over $\eta$ & Worst over $\eta$ \\
\midrule
1  & 95.98 & 96.13 & 95.82 \\
2  & 96.04 & 96.19 & 95.93 \\
4  & 96.04 & 96.21 & 95.79 \\
8  & 96.20 & 96.37 & 96.05 \\
12 & \textbf{96.42} & \textbf{96.59} & 96.29 \\
16 & 96.24 & 96.51 & 96.05 \\
20 & 96.24 & 96.44 & 96.12 \\
24 & 96.32 & 96.52 & 96.05 \\
28 & 96.33 & 96.51 & 96.19 \\
32 & 96.27 & 96.52 & 96.05 \\
\bottomrule
\end{tabular}
\caption{
Sensitivity to the number of selected image-attending heads $H_{\mathrm{top}}$.
For each $H_{\mathrm{top}}$, we report the mean, best, and worst Avg. \%Full over all tested energy ratios.
}
\label{tab:head_topk_sensitivity}
\end{table}

\subsection{Effect of Energy Ratio}
\begin{table*}[!t]
\centering
\small
\setlength{\tabcolsep}{6pt}
\begin{tabular}{cccc}
\toprule
Energy Ratio $\eta$ & Mean over $H_{\mathrm{top}}$ & Best over $H_{\mathrm{top}}$ & Worst over $H_{\mathrm{top}}$ \\
\midrule
0.05 & 96.15 & 96.42 & 95.94 \\
0.10 & 96.15 & 96.42 & 95.93 \\
0.20 & 96.12 & 96.31 & 95.90 \\
0.40 & 96.18 & 96.42 & 95.79 \\
0.60 & 96.23 & 96.49 & 96.05 \\
0.75 & 96.18 & \textbf{96.59} & 96.02 \\
0.85 & 96.22 & 96.48 & 96.02 \\
0.90 & 96.23 & 96.38 & 96.03 \\
0.95 & \textbf{96.29} & 96.47 & 96.06 \\
0.98 & 96.28 & 96.52 & 95.82 \\
0.99 & 96.25 & 96.51 & 95.88 \\
\bottomrule
\end{tabular}
\caption{
Sensitivity to the energy ratio $\eta$ used in the Tikhonov-regularized projection.
For each $\eta$, we report the mean, best, and worst Avg. \%Full over all tested values of $H_{\mathrm{top}}$.
}
\label{tab:energy_ratio_sensitivity}
\end{table*}
\begin{table*}[!t]
\centering
\setlength{\tabcolsep}{10pt}
\begin{tabular}{lccccc}
\toprule
Benchmark & Cov@22 & Cov@54 & Cov@92 & Cov@115 & Avg. \\
\midrule
MME     & 11.65 & 12.67 & 13.48 & 15.19 & 13.25 \\
TextVQA &  3.69 &  6.71 &  8.29 & 10.24 &  7.24 \\
POPE    & 14.06 & 13.08 & 14.40 & 15.27 & 14.20 \\
SQA     & 12.93 &  9.14 & 11.38 & 12.39 & 11.46 \\
\midrule
Average & 10.58 & 10.40 & 11.89 & 13.27 & 11.54 \\
\bottomrule
\end{tabular}
\caption{
Average overlap of visual token selections by attention and CMR across all 32 layers. All values are percentages.
}
\label{tab:attn_cmr_overlap}
\end{table*}
Table~\ref{tab:energy_ratio_sensitivity} summarizes the effect of the energy ratio $\eta$.
For each value of $\eta$, we report the mean, best, and worst average performance over all tested values of $H_{\mathrm{top}}$.
The mean performance varies only mildly across different energy ratios, showing that SIEVE is robust to the exact threshold used to determine the Tikhonov regularization scale.
Although the best single configuration appears at $\eta=0.75$, the overall mean performance remains stable across a wide range of values, confirming that the CMR computation does not rely on a carefully tuned energy threshold.

Overall, these results demonstrate that SIEVE is robust to both hyperparameters.
The number of selected image-attending heads has a slightly larger effect, where moderate head selection performs best.
In contrast, the energy ratio has only a mild influence, suggesting that the Tikhonov-regularized CMR computation is stable across a broad range of subspace energy thresholds.
This supports that the effectiveness of SIEVE mainly comes from cross-modal residual guidance and residual-space token selection, rather than from delicate hyperparameter tuning.

\section{Complementarity between Attention and CMR}
\label{app:attn_cmr_complementarity}

To verify that attention and CMR provide complementary rather than redundant token selection cues, we measure the overlap between the token subsets selected by the two scores before fusion. For a given layer and token budget $k$, let $\mathcal{S}^{\mathrm{attn}}_k$ and $\mathcal{S}^{\mathrm{CMR}}_k$ denote the top-$k$ visual token sets selected by the attention score and the CMR score, respectively. We define the overlap coverage as
\begin{equation}
\mathrm{Cov@}k =
\frac{
\left|\mathcal{S}^{\mathrm{attn}}_k \cap \mathcal{S}^{\mathrm{CMR}}_k\right|
}{k}.
\end{equation}
A lower $\mathrm{Cov@}k$ indicates that the two scores prefer more different token subsets.

Table~\ref{tab:attn_cmr_overlap} summarizes the average coverage over all 32 layers on four benchmarks. The overlap remains consistently low across different token budgets. The average coverage over all settings is only 11.54\%, and remains 13.27\% even under the larger top-115 budget. This suggests that attention and CMR are not redundant ranking signals. Attention mainly reflects query-conditioned semantic relevance, while CMR captures visual tokens with distinctive residual information that cannot be well explained by the text subspace.

This low overlap further explains why using either score alone is suboptimal. Attention-only selection may miss geometrically distinctive visual tokens with weak semantic saliency, while CMR-only selection may miss query-relevant tokens whose residual information is less prominent. Therefore, fusing the two scores covers a more complete set of useful visual evidence and leads to better performance than either single score. Meanwhile, multiplicative fusion is not equivalent to a hard intersection of two top-$k$ sets. Instead, it performs joint ranking over continuous scores: it retains tokens with both non-trivial semantic relevance and residual information, while suppressing background or noisy tokens whose support from either signal is close to zero. This also explains its advantage over additive fusion.

\section{Complexity Analysis and Efficiency Discussion}
\label{app:complexity}

\subsection{Complexity and Net Acceleration Condition}
\label{app:complexity_condition}

For an input sequence of length $n$, hidden dimension $d$, and FFN intermediate dimension $m$, the computational cost of a standard Transformer layer can be approximated as
\begin{equation}
F(n)=4nd^2+2n^2d+2ndm .
\end{equation}
Here, $4nd^2$ comes from the Q/K/V/O linear projections, $2n^2d$ comes from self-attention, and $2ndm$ comes from the FFN module. Therefore, reducing the sequence length decreases not only the linear projection and FFN costs, but also the quadratic self-attention cost.

If pruning reduces the sequence length from $n$ to $n'$, the per-layer computational saving is
\begin{equation}
\begin{aligned}
F(n)-F(n')
&=
(n-n')
\bigl[
4d^2+2d(n+n')  \\
&\quad
+2dm
\bigr].
\end{aligned}
\end{equation}
This shows that the saving depends on both the number of removed tokens, $n-n'$, and the original sequence scale, $n+n'$. Thus, token pruning becomes more beneficial when the input sequence is longer and visual tokens occupy a large portion of the sequence.

SIEVE introduces additional computation only at a few pruning layers. At a pruning layer, let $N_v$ be the number of visual tokens, $N_t$ be the number of post-image text tokens, and $k$ be the number of retained visual tokens. The scoring cost of SIEVE, including CMR projection and residual-space diversity selection, can be summarized as
\begin{equation}
\begin{aligned}
C_{\mathrm{score}}
=
O\bigl(
&N_t^2d + N_t^3 + N_vN_td \\
&+ N_vN_t^2 + N_v^2d + kN_v
\bigr).
\end{aligned}
\end{equation}
The first two terms come from text-subspace construction and Tikhonov-regularized solving, the next two terms come from projecting visual tokens onto the text subspace, and the last two terms come from residual-space similarity computation and greedy diversity selection. Since $N_t$ is usually much smaller than $N_v$ and $d$, the dominant cost is typically the residual-space Gram matrix:
\begin{equation}
C_{\mathrm{score}} \approx O(N_v^2d).
\end{equation}

Although SIEVE introduces this scoring overhead, it follows a ``one-time scoring, multi-layer saving'' pattern. The scoring is performed only at pruning layers, while the shortened sequence benefits all subsequent Transformer layers. Let $n_0$ be the original sequence length, $n_l$ be the actual sequence length at layer $l$, and $\mathcal{P}$ be the set of pruning layers. SIEVE achieves net acceleration when
\begin{equation}
\sum_l
\left[
F(n_0)-F(n_l)
\right]
>
\sum_{p\in\mathcal{P}}
C_{\mathrm{score}}^p .
\end{equation}
That is, SIEVE improves inference efficiency when the accumulated saving from shorter sequences in later layers exceeds the scoring overhead introduced at pruning layers. This condition also explains why SIEVE is more effective when pruning is performed earlier, more tokens are removed, and the original sequence is longer.

\subsection{Discussion on Long-Sequence Inputs}
\label{app:long_sequence_discussion}

For long-sequence inputs, the efficiency behavior depends on where the sequence length comes from. The first and most relevant case is long visual sequences, such as high-resolution images, multi-image inputs, or videos. In these scenarios, the number of visual tokens $N_v$ increases, which also increases the scoring cost of SIEVE, especially the $O(N_v^2d)$ residual-space similarity computation. However, standard Transformer self-attention also contains the $O(n^2d)$ term, and this cost is paid at every subsequent layer. Since SIEVE scoring is only performed at a few pruning layers, while the benefit of shorter sequences accumulates across many later layers, the total computational saving can outweigh the additional scoring cost when pruning is applied early.

The second case is long text prompts. If the post-image text length $N_t$ becomes very large, the cost of text-subspace construction and Tikhonov projection may increase. In this case, one can use a fixed-length post-image text window or restrict the effective rank of the text subspace, so that the terms related to $N_t$ remain controlled.

Overall, SIEVE is especially suitable for long visual-sequence scenarios where visual tokens dominate the computational cost. In such cases, the original Transformer cost grows rapidly with sequence length, while SIEVE only introduces limited scoring overhead at a few pruning layers. The reduced token sequence is then reused by subsequent layers, leading to lower overall inference complexity and improved efficiency.

\label{sec:appendix}

\end{document}